\documentclass[default,iicol]{sn-jnl}

\usepackage{natbib}
\usepackage{graphicx}%
\usepackage{multirow}%
\usepackage{amsmath,amssymb,amsfonts}%
\usepackage{amsthm}%
\usepackage{mathrsfs}%
\usepackage[title]{appendix}%
\usepackage{xcolor}%
\usepackage{textcomp}%
\usepackage{manyfoot}%
\usepackage{booktabs}%
\usepackage{algorithm}%
\usepackage{algorithmicx}%
\usepackage{algpseudocode}%
\usepackage{amssymb}
\usepackage{bbding}
\usepackage{hyperref}

\begin{document}

\title[Article Title]{Breathing Life into Faces: Speech-driven 3D Facial Animation with Natural Head Pose and Detailed Shape}


\author[1]{\fnm{Wei} \sur{Zhao}}\email{zhaoweiheap@hnu.edu.cn}
\equalcont{These authors contributed equally to this work.}

\author[1]{\fnm{Yijun} \sur{Wang}}\email{wyjun@hnu.edu.cn}
\equalcont{These authors contributed equally to this work.}

\author[2]{\fnm{Tianyu} \sur{He}}\email{tianyuhe@microsoft.com}

\author[1]{\fnm{Lianying} \sur{Yin}}\email{yin2110@hnu.edu.cn}

\author*[1]{\fnm{Jianxin} \sur{Lin}}\email{linjianxin@hnu.edu.cn}

\author*[3]{\fnm{Xin} \sur{Jin}}\email{jinxin@eias.ac.cn}

\affil*[1]{\orgdiv{The college of Computer Science and Electronic Engineering}, \orgname{Hunan University}, \orgaddress{ \city{Hunan}, \country{China}}}

\affil[2]{\orgname{Microsoft Research Asia}, \orgaddress{ \city{Beijing}, \country{China}}}

\affil[3]{ \orgname{Eastern Institute of Techonology}, \orgaddress{ \city{Ningbo}, \country{China}}}


\abstract{
The creation of lifelike speech-driven 3D facial animation requires a natural and precise synchronization between audio input and facial expressions. However, existing works still fail to render shapes with flexible head poses and natural facial details (e.g., wrinkles). This limitation is mainly due to two aspects: 1) Collecting training set with detailed 3D facial shapes is highly expensive. This scarcity of detailed shape annotations hinders the training of models with expressive facial animation. 2)  Compared to mouth movement, the head pose is much less correlated to speech content. Consequently, concurrent modeling of both mouth movement and head pose yields the lack of facial movement controllability. To address these challenges, we introduce VividTalker, a new framework designed to facilitate speech-driven 3D facial animation characterized by flexible head pose and natural facial details. Specifically, we explicitly disentangle facial animation into head pose and mouth movement and encode them separately into discrete latent spaces. Then, these attributes are generated through an autoregressive process leveraging a window-based Transformer architecture. To augment the richness of 3D facial animation, we construct a new 3D dataset with detailed shapes and learn to synthesize facial details in line with speech content. Extensive quantitative and qualitative experiments demonstrate that VividTalker outperforms state-of-the-art methods, resulting in vivid and realistic speech-driven 3D facial animation. 
\href{https://weizhaomolecules.github.io/VividTalker/}{https://weizhaomolecules.github.io/VividTalker/}
}

\keywords{3D Facial Animation,  Detailed Face Shape,  Motion Disentanglement}



\maketitle

\section{Introduction}\label{sec1}

The domain of 3D virtual facial animation has gained significant attention and research interest in both academic and industrial sectors due to its immense value across domains like entertainment, communication, and healthcare. Success in 3D virtual facial animation relies on exhibiting human-like characteristics, including synchronization and naturalness. Synchronization involves creating believable animations that are aligned with user expectations, bridging the gap between virtual avatars and the real world. Naturalness focuses on capturing human behavior, movements, and expressions to evoke emotional connections and user engagement.


Recent research has made significant strides in enhancing the synchronization of 3D avatar animation, particularly the development of data-driven approaches, which learn the mapping from speech audio to facial motions in an end-to-end manner \cite{cudeiro2019capture, richard2021meshtalk, fan2022faceformer, xing2023codetalker, peng2023emotalk, thambiraja2022imitator, ezzat2000visual}. Cudeiro et al.~\cite{cudeiro2019capture} is able to generate lip animation and upper facial movements from the audio input with the ability of generalization across different identities. However, such an approach normally exhibits mild or static upper face animation, especially in facial areas that exhibit minimal or no correlation with the audio signal. MeshTalk \cite{richard2021meshtalk} enables more realistic motion synthesis for the entire face, where they disentangle audio-correlated and audio-uncorrelated information in a categorical latent space. The categorical latent space is depicted using number values rather than feature vectors, leading to challenging training and consequently hindering its performance. CodeTalker \cite{xing2023codetalker} employed a codebook-based mechanism to further address potential issues with average results. On the other hand, FaceFormer \cite{fan2022faceformer} adopted a transformer model with pre-trained speech representations to predict facial motions to address the long-term context issues. One of our observations is that the 3D facial model contains a highly decoupled representation, i.e., mouth movement and head pose correlate differently to speech content. However, few of these methods have taken the concurrent modeling of both mouth movement and head pose into consideration, yielding the lack of facial movement controllability.


Existing data-driven 3D facial animation methods also rely on high-quality 3D audio-visual data for model training. However, high-quality vision-based motion capture of the user is highly time-consuming and resource-consuming. In addition, the existing face animation datasets (e.g, VOCASET \cite{cudeiro2019capture}  and BIWI dataset \cite{fanelli20103}) typically lack detailed shape information (e.g., wrinkles) and head poses, making it hard to train the model with enough natural facial animation features. Besides, the existing 2D talking head datasets ( e.g., LRS3-TED \cite{afouras2018lrs3}, HDTF \cite{zhang2021flow}) only provide the raw data like audio files and video files without 3D facial coefficients. As shown in Table \ref{tab:dataset},  current audio-visual datasets lack either detailed shape information or realistic 3D facial coefficients.




To address the above challenges and advance the progress of vivid 3D facial animation, we initially propose a new high-resolution 3D audio-face dataset from online sources, which we named the 3D Vivid Talking Face Dataset (3D-VTFSET). This dataset was sourced from youtube website, covering $300+$ subjects, and comprises approximately $20+$ hours of videos. Compared to existing datasets, our 3D-VTFSET dataset contains 3D head from a larger pool of in-the-wild subjects with detailed face geometry, thus enabling more detailed expression capture and animation.

\begin{table*}[]
\centering
\caption{ \centering{Statistics of current 3D facial datasets.}}
\label{tab:dataset}\footnotesize
\begin{tabular}{c|l|l|l|c|c|c|c}
\hline
Dataset Name & \#Subjects & \#Sentences & \multicolumn{1}{c|}{\#Hours} & Mouth Movement            & Detailed Shape & Head Pose & In-the-Wild\\ \hline
BIWI \cite{fanelli20103}         & 14       & 40        & 0.76                          & \Checkmark & \XSolidBrush              & \XSolidBrush       & \XSolidBrush   \\ \hline
VOCASET  \cite{cudeiro2019capture}    & 12       & 255       & 2.58                           & \Checkmark                         & \XSolidBrush              & \XSolidBrush     & \XSolidBrush    \\ \hline
S3DFM   \cite{zhang20193d}     & 100      & 199       & 16.5                           & \Checkmark                          & \XSolidBrush            & \XSolidBrush        & \XSolidBrush  \\ \hline
Multiface  \cite{wuu2022multiface}  & 13       & 50        & 0.9                           & \Checkmark                          & \XSolidBrush              & \XSolidBrush        & \XSolidBrush  \\ \hline
3D-VTFSET (Ours)         & \textbf{300+}     & \textbf{10K+}      & \multicolumn{1}{l|}{\textbf{20.8}}    & \Checkmark                          & \Checkmark             & \Checkmark & \Checkmark          \\ \hline

\end{tabular}
\end{table*}

Next, we introduce VividTalker, an innovative framework for vivid speech-driven 3D head animation, which is specifically designed to handle aforementioned challenges via \textit{factor disentanglement} and \textit{detail enrichment} without relying on 3D facial scans. Specifically, to effectively address the issue of feature learning conflict, we first employ two VQ-VAE models~\cite{van2017neural} to encode the head pose and mouth movement into discrete latent spaces separately. Then, to predict motion dynamics over the learned discrete latent space, we employ a window-based Transformer for autoregressive motion prediction. This network architecture enables us to generate highly accurate and realistic motion sequences for the dynamic head, enhancing the overall quality and fidelity. Furthermore, for enriching sophisticated details in facial animation, we leverage a pre-trained DECA model~\cite{feng2021learning}, which estimates the coefficients of 3D Morphable Models (3DMMs)~\cite{blanz1999morphable} as well as detailed shape, to construct rich training set. In this way, our window-based Transformer is trained to predict the disentangled coefficients and detailed shape simultaneously.

The main contributions of our work are as follows:
\begin{itemize}
    \item We build a large-scale high-resolution 3D audio-face dataset with detailed shape information and flexible head pose, namely 3D-VTFSET, using an in-the-wild 3D face reconstruction model. We also apply a data smoothing method to reduce noise and fluctuation introduced by the reconstruction model. 

    \item We propose a new method for vivid speech-driven 3D head animation, namely VividTalker, to explicitly disentangle facial animation into head pose and mouth movement and encode them using two separate codebooks.

    \item To the best of our knowledge, we are the first to enrich the 3D head animation by predicting dynamic detailed shapes directly from speech signals using a window-based Transformer, enhancing the overall visual fidelity.

\end{itemize}

Extensive experimental results show that our method achieves state-of-the-art performance, showing a vivid and realistic speech-driven 3D facial animation effect. 

\section{Related Work}\label{relatework}

\subsection{2D Talking Head Animation}
There has been a significant interest in facial animation in recent years, particularly in the 2D-based facial animation \cite{chen2020talking,chen2018lip,chiu2019action,chung2017out,du2019bio,das2020speech,fan2015photo,ji2021audio,prajwal2020lip,liu2021geometry,vougioukas2020realistic}. 

Early efforts in talking head generation initially focused on creating realistic mouth movements for specific individuals. Suwajanakorn \textit{et al.}  \cite{suwajanakorn2017synthesizing} produced a high-quality video of Barack Obama with precise lip synchronization to input audio, integrating these synthesized lip motions into a target video clip. While this method yielded excellent talking head videos, its limited versatility restricted its broader application. In contrast, the approach presented in Prajwal et al. \cite{prajwal2020lip} not only refined a pretrained discriminator to enhance lip-sync accuracy but also facilitated the generation of talking face videos featuring arbitrary identities using a target identity encoder. Some other methods, such as ATVG \cite{chen2019hierarchical}, and MakeItTalk \cite{zhou2020makelttalk}, utilized facial landmarks as an intermediary step to guide video generation. These methods decoupled content and identity information from the input audio, attempting to extract invariant features and style features from the audio. Additionally, many works in this domain employed 3DMM coefficients as an intermediary representation for generation \cite{karras2017audio,zhang2022sadtalker,doukas2021headgan,yin2022styleheat}.

After addressing the issue of lip-sync accuracy, researchers have shifted their focus to the head pose \cite{ackland2019real, kang2023expression, chen2022learning}. Like lip-sync accuracy, head pose directly affects the realism of the generated videos \cite{zhou2021pose,jang2023s,lu2021live}. 
Due to the lack of strong one-to-one correspondence between the audio signal and head pose, most works chose to use a specific pose as a constraint for video synthesis \cite{zhou2021pose, jang2023s, pumarola2020ganimation, ververas2020slidergan, garg2023visually}. Some works had also begun to explore whether there was a correspondence between head pose and audio, which might not be unique \cite{zhang2022sadtalker, lu2021live, zhang2021facial}. Unlike traditional methods that focused solely on lip synchronization, FACIAL \cite{zhang2021facial} incorporated phonetics-aware, context-aware, and identity-aware information to generate natural and convincing talking faces.

While head pose has shown some promising results in 2D talking face animation, it has received little attention in the field of 3D head animation. Therefore, we design a VQ-VAE-based method to reconstruct the 3D head pose for vivid generation in this work.

\subsection{Speech-driven 3D Head Animation}
Similar to the development process of 2D facial animation,  early research on 3D facial animation initially focused on mouth synchronization \cite{edwards2016jali, taylor2012dynamic, xu2013practical, cohen2001animated}. Traditional approaches, which segmented speech into phonemes and assigned corresponding visemes, faced limitations due to overlapping mouth movements and context-dependent variations in mouth movements for the same pronunciation. The viseme-based method \cite{taylor2012dynamic} generated realistic speech visual animations by mapping phonemes to dynamic mouth movements. JALI \cite{edwards2016jali} incorporated psycholinguistic knowledge to identify jaw and lip movement patterns, mapping phonemes to a multi-valued visual representation. However, these approaches often require significant manual effort, particularly from artists, to adjust related parameters for the final animations.

More recent methods proposed end-to-end learning approaches for facial animation modeling \cite{karras2017audio, cudeiro2019capture, richard2021meshtalk, fan2022faceformer}.
VisemeNet~\cite{zhou2018visemenet} utilized a three-stage LSTM network to generate lip motion aligned with speech style.
Karras \textit{et al.}~\cite{karras2017audio} directly converted audio signals into 3D vertex coordinates with the emotional state using machine learning techniques.
VOCA~\cite{cudeiro2019capture} introduced style encoding to account for speaker-dependent speaking styles, achieving fine mouth synchronization but focusing on the lower face region.
Meshtalk~\cite{richard2021meshtalk} incorporated a cross-modal loss to disentangle audio-correlated and audio-uncorrelated features, enabling plausible upper-face animation.
To mitigate occasional jitters caused by short-term audio windows such as VOCA and MeshTalk, Faceformer~\cite{fan2022faceformer} employed a transformer-based method for autoregressively predicting long-term 3D motion sequences.
However, none of these methods considers head pose and detailed shape. In this paper, we propose a factor disentanglement mechanism to handle a head pose issue and a window-based Transformer mechanism to address the detailed shape for improving the vividness of generation.

\begin{figure*}
    \centering
    \includegraphics[width=14cm]{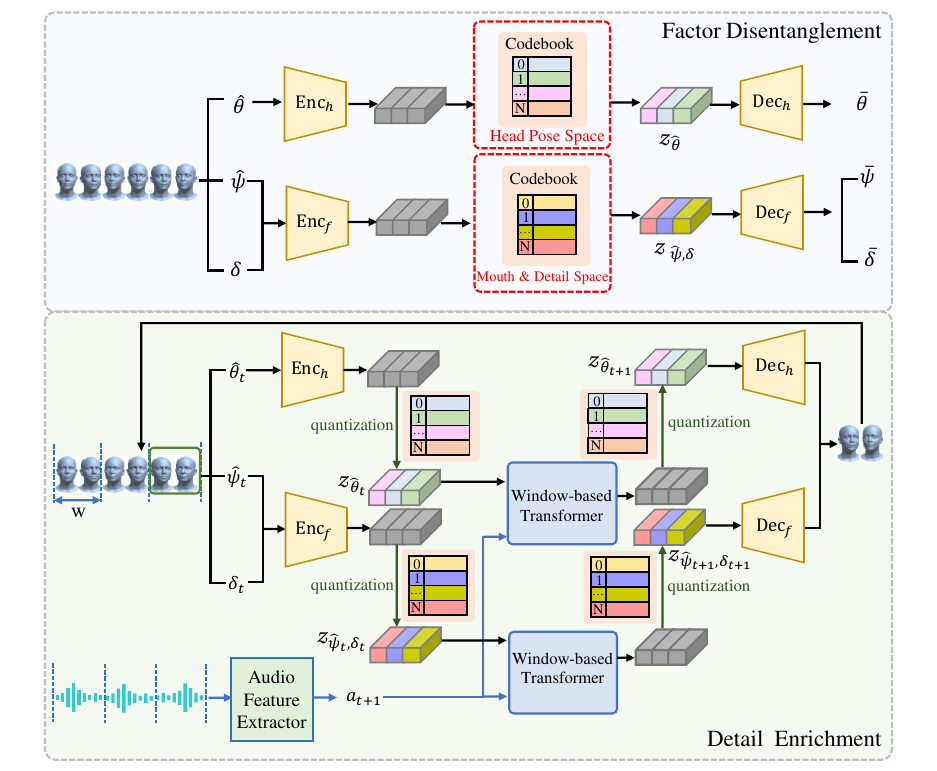}
    \caption{Overall pipeline of the proposed VividTalker. Our method is composed of two core components: 1) the factor disentanglement module utilizes two VQ-VAE models to encode the head pose and mouth movement into separate discrete latent spaces; 2) the detail enrichment module employs a window-based Transformer to predict motion dynamics (including facial details) over the learned discrete latent space, given an audio signal.}
    \label{architecture}
\end{figure*}

\section{VividTalker}

\subsection{Data Construction}
\label{dataConstruction}

Previous works on 3D head animation mainly focused on the VOCA dataset~\cite{cudeiro2019capture}. However, this dataset lacks diversity in terms of variations in head pose and detailed shape and mainly focuses on mouth movements. When targeting generating realistic 3D head animations, relying solely on mouth movements and slight facial expressions is insufficient. Since the large-scale and precise 3D data is difficult to acquire, we construct a new dataset by the pre-trained DECA model~\cite{feng2021learning}, which can extract head pose, facial expression, and detailed shape from the input image. DECA~\cite{feng2021learning} adopts FLAME~\cite{li2017learning} as 3D Morphable Models (3DMMs)~\cite{blanz1999morphable}, which utilizes corrective coefficients with $N=5,023$ vertices and incorporates $4$ joints representing the neck, jaw, and eyeballs. FLAME~\cite{li2017learning} can be characterized by the following function:
\begin{equation}
   M(\vec{\beta}, \vec{\theta}, \vec{\psi}):\mathbb{R}^{|\vec{\beta}| \times|\vec{\theta}| \times|\vec{\psi}| }\rightarrow \mathbb{R}^{3N},
\end{equation}
where $\vec{\beta}$ denotes shape coefficients, $\vec{\theta}$ denotes pose coefficients and $\vec{\psi}$ denotes expression coefficients. Given a template mesh $\overline{\mathrm{T}}$ and shape, pose, and expression blendshapes and coefficients, FLAME rotates the vertices to $\overline{\mathrm{T}}$ with deformation of pose $\vec{\theta}$ and expression $\vec{\psi}$:
\begin{equation}
T_{P}(\vec{\beta}, \vec{\theta}, \vec{\psi})=\overline{\mathbf{T}}+B_{S}(\vec{\beta} ; \mathcal{S})+B_{P}(\vec{\theta} ; \mathcal{P})+B_{E}(\vec{\psi} ; \mathcal{E}).
\end{equation}
For detailed shape, we use geometric displacements $\delta$, a 128-dimensional latent code extracted by DECA~\cite{feng2021learning} to enrich 3D head animation. The latent code $\delta$ is decoded into displacement map $\mathcal{D}$ along with $\psi$ and $\theta$ by an decoder designed by DECA~\cite{feng2021learning}:
\begin{equation}
    \mathcal{M_{\delta}} =Dec_{\delta}(\delta, \psi, \theta).
\end{equation}
We directly use the pre-trained $Dec_{\delta}$ to obtain the displacement map $\mathcal{M_{\delta}}$.

\subsection{Our Method}


As shown in Figure~\ref{architecture}, we explicitly disentangle facial animation into head pose and mouth movement in discrete latent spaces and encode them using two separate codebooks. By utilizing the separated codebooks, the space of slighter movement but more correlated to the audio, i.e., mouth movement and detailed shape, and more drastic movement but less correlated to the audio, i.e., head pose, can both be efficiently represented in a discrete format.


\subsubsection{Factor Disentanglement with Two Separate VQ-VAEs}
\label{VQ-VAE}


Human mouth movement is mainly driven by the speech content while the head pose is more likely to be affected by others such as personality, habit, mood, etc. Motivated by this, we propose to model the head pose and expression coefficients separately. Since the detailed shape is synchronized with the mouth movement, the latent code $\delta$ of the detailed shape is concatenated to the expression coefficients and jointly modeled.

As shown in the upper part of Figure \ref{architecture}, specifically, we learn two separate representations through two autoencoder networks with vector quantization in the bottleneck (VQ-VAE)~\cite{van2017neural}. 
We split the pose coefficient $\theta$ mentioned in Section \ref{dataConstruction} into two parts, as the first $3$ dimensions are related to the head pose while the last $3$ dimensions are related to mouth movement. Therefore, we use $\hat{\theta}$ to denote head pose, which is the first $3$ dimension in $\theta$. And the mouth movement is denoted as $\hat{\psi}$, which contains the last $3$ dimensions in $\theta$ and expression coefficients $\psi$.
The first VQ-VAE (parameterized with $\text{Enc}_{h}$ and $\text{Dec}_{h}$) is responsible for obtaining the discrete latent codes of head pose $\hat{\theta}$. 
While the second VQ-VAE (parameterized with $\text{Enc}_{f}$ and $\text{Dec}_{f}$) aims at obtaining the discrete latent codes of mouth movement $\hat{\psi}$ and detailed shape $\delta$. Formally, given motion input $\hat{\theta}_{t}$, $\hat{\psi}_{t}$ and $\delta_{t}$ at time stamp $t$, they are separately encoded by two encoders $\text{Enc}_{h}$ and $\text{Enc}_{f}$ into two latent codes. Then, we obtain the discrete head pose latent $z_{\hat{\theta}}$, mouth movements and detailed shape latent $z_{\hat{\psi},\delta}$ by a quantization operator $\mathbf{quant()}$:
\begin{equation}
\label{encodeH}
  z_{\hat{\theta}_{t}} = \mathbf{quant}(\hat{\theta}_{t}):=\underset{e_{i} \in \mathcal{E}_h}{\arg \min }\left\|\text{Enc}_{h}(\hat{\theta_t})-e_{i}\right\|_{2},
\end{equation}
\begin{equation}
\label{encodeM}
\begin{split}     z_{\hat{\psi}_{t},\delta_{t}} &= \mathbf{quant}(\hat{\psi}_{t}, \delta_t)\\
&:=\underset{e_{j} \in \mathcal{E}_f}{\arg \min }\left\|\text{Enc}_{f}(\text{Concat}(\hat{\psi}_{t}, \delta_t)-e_{j}\right\|_{2},
\end{split}
\end{equation}
where $e_{i} $, $e_{j}$ denotes a latent embedding vector from latent embedding space $\mathcal{E}_h$, $\mathcal{E}_f$ respectively.
We can obtain reconstruction $\hat{\theta}_{t}$, $\hat{\psi}_{t}$ and $\delta_{t}$ through $\text{Dec}_{h}$ and $\text{Dec}_{f}$ :
\begin{equation}
    \bar{\theta}_{t}=\text{Dec}_h(z_{\hat{\theta}_{t}}), \quad 
 \bar{\psi}_{t},\bar{\delta}_t=\text{Dec}_f(z_{\hat{\psi}_{t},\delta_{t}}).
\end{equation}

During the training stage, the two VQ-VAE models are trained to map the head pose, mouth movements, and detailed shape into latents with minimum reconstruction error.



\subsubsection{Detail Enrichment with A Window-based Transformer}
\label{prediction}
In this module, firstly, the raw head pose $\hat{\theta}_{1:T} $, mouth movements $\hat{\psi}_{1:T} $ and dynamic details $\delta_{1:T}$ will be splitted into $T^* = T/w$ time stamps. Each window contains $w$ frames, where $w$ is a window length that satisfies $1<w<T$. 
Then, we split the audio signal accordingly and extracted the mel spectrum feature of it, resulting in audio feature $\{a_{t}\}_{t=0}^{T^*}$. We initialize  $\hat{\theta}_{0}$, $\hat{\psi}_{0}$ and $\delta_{0}$ to 0 at the beginning.

Secondly, each $\hat{\theta}_{t}$, $\hat{\psi}_{t}$, $\delta_{t}$ will be encoded into latent code  $z_{\hat{\theta}_{t}}$,  $z_{\hat{\psi}_{t},\delta_{t}}$ according to the by pre-trained $\text{Enc}_{h}$ and 
 $\text{Enc}_{f}$ according to Equation~\ref{encodeH} and \ref{encodeM}.
Instead of using causal attention like previous works~\cite{fan2022faceformer, xing2023codetalker}, we implement a full-attention mechanism within each window. Firstly, we concatenate the audio features $a_{t+1}$ and quantized results of head motion, mouth movement, and detailed shape separately for the construction of $Q, K, V$ matrices as follows:
\begin{equation}
\begin{split}
Q_{a_{t+1},\hat{\theta}_{t}} =& \text{Concat}(a_{t+1}, z_{\hat{\theta}_{t}})\times W_{Q},\\
Q_{a_{t+1},\hat{\psi}_{t},\delta_{t}} =& \text{Concat}(a_{t+1}, z_{\hat{\psi}_{t},\delta_{t}}) \times W_{Q},\\
\end{split}
\end{equation}
\begin{equation}
\begin{split}   K_{a_{t+1},\hat{\theta}_{t}} =& \text{Concat}(a_{t+1}, z_{\hat{\theta}_{t}})\times W_{K},\\
K_{a_{t+1},\hat{\psi}_{t},\delta_{t}} =& \text{Concat}(a_{t+1}, z_{\hat{\psi}_{t},\delta_{t}}) \times W_{K},\\
\end{split}
\end{equation}
\begin{equation}
\begin{split} 
V_{a_{t+1},\hat{\theta}_{t}} =& \text{Concat}(a_{t+1}, z_{\hat{\theta}_{t}})\times W_{V},\\
V_{a_{t+1},\hat{\psi}_{t},\delta_{t}} =& \text{Concat}(a_{t+1}, z_{\hat{\psi}_{t},\delta_{t}}) \times W_{V},\\
\end{split}
\end{equation}
where $z_{\hat{\theta}_{t}}$ denotes the quantization of head pose while $z_{\hat{\psi}_{t},\delta_{t}}$ denotes the quantization of detailed shape and mouth movement. Then, attention is used to calculate the probability of the head pose, mouth movement, and details.
\begin{equation}
   z_{\hat{\theta}_{t+1}} = \text{S}(\frac{Q_{a_{t+1},\hat{\theta}_{t}}\times K_{a_{t+1},\hat{\theta}_{t}}+M}{{\sqrt{D}}})V_{a_{t+1},\hat{\theta}_{t}},
\end{equation}

\begin{equation}
\begin{aligned}
   z_{\hat{\psi}_{t+1},\delta_{t+1}} &= \text{S}(\frac{Q_{a_{t+1},\hat{\psi}_{t},\delta _t}\times K_{a_{t+1},{\psi}_{t},\delta _t} + M}{{\sqrt{D}}})\\  &\cdot V_{a_{t+1},\hat{\psi}_{t},\delta _t},    
\end{aligned}
\end{equation}

where $S$ denotes $softmax$ operation and $M$ denotes $Mask$ designed for preventing the model from calculating self-attention on the audio token. The output features $z_{\hat{\theta}_{t+1}}$ and $z_{\hat{\psi}_{t+1},\delta_{t+1}}$ are further decoded by the pre-trained VQ-VAE decoders to produce the motion prediction. By incorporating cross-modality information fusion, our method achieves the ability to accurately reconstruct facial details solely from audio signals. This represents a significant advancement in the field of 3D head animation, as no previous approach has been proposed to obtain such detailed reconstruction solely from audio inputs.


During the training stage, our window-based Transformer architecture learns to predict multiple future motions within a window, unlike previous auto-regressive methods (only a single frame prediction),  which can capture contextual information in the historical window effectively. The training process employs the teacher-forcing mode that takes real sample quantization as past motion input. In the testing phase, an auto-regressive mechanism predicts only the current motion. After each prediction step, the window shifts forward, encompassing the next $w+1$ time steps of input features. This sliding window approach repeats until all time steps are processed.


\subsubsection{Training Objective}
\label{loss}

We use reconstruction loss $\mathcal{L}_{\mathrm{rec}}$ and commitment loss $\mathcal{L}_{\mathrm{cmt}}$ to supervise the training of each VQ-VAE model:
\begin{equation}
\begin{split}
   \mathcal{L}_{\mathrm{\hat{\theta}_t }} &= \mathcal{L}_{\mathrm{rec_{\hat{\theta }_t }}} + \mathcal{L}_{\mathrm{cmt_{\hat{\theta}_{t} }}} \\
    &=\|\overline{\theta }_t-\hat{\theta }_t\|  + \left\|\text{sg}[\text{Enc}_{h}(\hat{\theta }_t)]-e_i\right\|_2^2 \\
    &+\beta \left\|\text{sg}\left[e_i\right]-\text{Enc}_{h}(\hat{\theta }_t)\right\|_2^2,
\end{split}
\end{equation}

\begin{equation}
\begin{split}
    \mathcal{L}_{\mathrm{\hat{\psi}_t },\delta_t} &= \mathcal{L}_{\mathrm{rec_{\hat{\psi}_t,\delta _t }}} + \mathcal{L}_{\mathrm{cmt_{\hat{\psi}_t,\delta _t }}} \\
    &=(\|\overline{\psi }_t-\hat{\psi }_t\|  +\|\overline{\delta}_t-\hat{\delta  }_t\|) \\
&+ \left\|\text{sg}[\text{Enc}_{f}(\text{Concat}(\hat{\psi}_t,\delta _t ))]-e_j\right\|_2^2 \\
&+\beta \left\|\text{sg}\left[e_j\right]-\text{Enc}_{f}(\text{Concat}(\hat{\psi}_t,\delta _t ))\right\|_2^2,
\end{split}
\end{equation}
where $\text{sg}$ stands for the stop-gradient operation, $\beta$ denotes the weight factor controlling the update rate between the codebook and encoder, $\|\overline{\theta }_t-\hat{\theta }_t\| $, $|\overline{\psi }_t-\hat{\psi }_t\|$, $\|\overline{\delta}_t-\hat{\delta  }_t\| $ is used to train encoder $Enc_{f}$, $Enc_{h}$ and decoder $Dec_{f}$, $Dec_{h}$. The codebook $\mathcal{E}_h$, $\mathcal{E}_f$ are learned by the second term that forces the embedding vectors $e_i$, $e_j$ towards the encoder outputs $\text{sg}[\text{Enc}_{h}(\hat{\theta }_t)]$, $\text{sg}[\text{Enc}_{f}(\text{Concat}(\hat{\psi}_t,\delta _t ))]$. The third term is used for encouraging the output of encoder $\text{Enc}_{h}(\hat{\theta }_t)$, $\text{Enc}_{f}(\text{Concat}(\hat{\psi}_t,\delta _t ))$ to stay close to the codebook vector $e_i$, $e_j$.

After training VQ-VAE, we focus on training the Transformer for cross-modality mapping while keeping the codebooks and motion encoder-decoder frozen to maintain their previous state.
We use mean square error on predicted head pose and mouth movement coefficients to guide the Transformer learning process:
\begin{align}
    \mathcal{L}_{\mathrm{MSE_{\hat{\theta }_t }}} &= \|\hat{\theta}_{t} -\overline{\theta}_{t} \|_2^2+\|z_{\hat{\theta }_{t} }-z_{\overline{\theta}_{t}}\|_2^2,
\end{align}

\begin{equation}  
\begin{split} \mathcal{L}_{\mathrm{MSE_{\hat{\psi}_t,\delta _{t}}}} &= \|\hat{\psi}_{t} -\overline{\psi}_{t} \|_2^2+\|\delta_{t} -\overline{\delta}_{t} \|_2^2\\
&+\|z_{\hat{\psi }_{t},\delta _t }-z_{\overline{\psi}_{t},\overline{\delta} _t}\|_2^2.
\end{split}
\end{equation}

\section{Experiments}
\label{experiments}
\subsection{Implementation Details}
The shape of raw input for $Enc_{h}$ is $T\times C_{\hat{\theta}}$ where $C_{\hat{\theta}} = 3$  while the shape of raw input for $Enc_{f}$ is $T\times C_{\hat{\psi}, \delta}$ where $C_{\hat{\psi}, \delta} = 181 $. Both encoder $Enc_{h}$ and $Enc_{f}$ are composed of $12$ Transformer encoder hidden layers with $8$ attention heads. The codebook size is set to $K \times N_{e}$, where  $K = 256$, $N_{e} = 256$. Both decoder $Dec_{h}$ and $Dec_{f}$ are composed of $12$ Transformer hidden layers and $8$ attention heads. For window-based Transformer, the window size is set to 12. Our model is implemented using PyTorch with Adam optimizer, employing a learning rate of $1e-2$. Training processes are performed on two Nvidia A100 GPUs with a batch size of 64. Our model takes approximately three days for VQ-VAE training and one day for the window-based Transformer training. The learning rate decays into $1e-6$ when the epoch reaches 400.


We provide \textbf{supplementary videos} that correspond to the results of our method and other state-of-the-art methods MeshTalk \cite{richard2021meshtalk}, FaceFormer \cite{fan2022faceformer}, CodeTalker \cite{xing2023codetalker}  and SadTalker \cite{zhang2021facial} as comparisons.

\begin{table*}[t]
\caption{\centering{Quantitative comparison with other methods on 3D-VTFSET dataset.}}
\centering
\footnotesize

    \begin{tabular}{lccccccc}
        \hline
           & Pose Error $\downarrow$ & Mouth Error $\downarrow$ & Detail Error $\downarrow$ & FD  $\downarrow$            & Diversity $\uparrow $    & LSE-C   $\uparrow $        & LSE-D  $\downarrow$          \\ \hline
SadTalker \cite{zhang2022sadtalker}  & -               & -                    & -                    & -              & 1.031               & 0.964               &  13.174               \\
MeshTalk \cite{richard2021meshtalk}   & 22.009          & 69.663               & -              & 0.093          & 0.708          & 0.386          & 14.124          \\
FaceFormer \cite{fan2022faceformer} & 19.228          & 57.147               & -               & 0.054          & 1.187          & 0.496          & 13.397          \\
CodeTalker \cite{xing2023codetalker} & 15.236          & 29.060               & -               & 0.050          & 1.045          & 0.833          & 13.762          \\
Ours       & \textbf{8.852}  & \textbf{22.895}      & \textbf{46.984}      & \textbf{0.027} & \textbf{1.239} & \textbf{1.275} & \textbf{12.469} \\ \hline
    \end{tabular}

\label{tab:comparison}
\end{table*}

\begin{figure*}[t]
    \centering
    
    \includegraphics[width=0.9\textwidth]{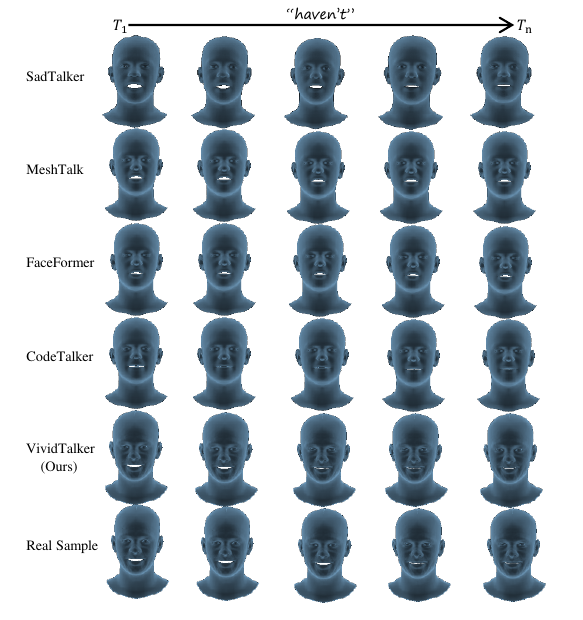}
    \caption{Visulization results. Compared to other methods, VividTalker(Ours) presents more consistent head pose variations with Real Sample and exhibits more accurate mouth movements (Please zoom in for better viewing). }
    \label{figcomparison}
\end{figure*}

\subsection{Dataset}
\paragraph{Data Collection} We use DECA \cite{cudeiro2019capture} for automatically generating the 3D talking dataset for 3D facial animation task. The dataset consists of over 20 hours of video, containing over 300 subjects from speech videos in English, downloaded from YouTube. The facial extraction is processed with a resolution of $512 \times 512$ on a frame rate of $30 fps$. The audio tracks are processed with a sample rate of 16kHz. Our dataset is organized into three sets: $train$, $val$ and $test$. Those three sets are completely independent. As indicated in Table \ref{tab:dataset}, our dataset significantly surpasses existing datasets such as BIWI \cite{fanelli20103}, VOCASET \cite{cudeiro2019capture}, S3DFM \cite{zhang20193d}, and Multiface \cite{wuu2022multiface}, featuring an extensive collection of 300+ subjects. In comparison, the second-largest dataset, S3DFM \cite{zhang20193d}, consists of only 100 subjects. Moreover, our dataset boasts a substantial text corpus, containing over 10,000 sentences, while other datasets typically offer a maximum of 255 sentences. The incorporation of our processing pipeline also simplifies the task of dataset expansion. 


\paragraph{Data Processing}
After extracting coefficients with DECA \cite{cudeiro2019capture}, we proceed to re-render sequences for assessing dataset quality. Regrettably, the raw output from DECA falls short of practical implementation standards, exhibiting occasional jitters in the rendered results. These fluctuations can introduce instability in smoothing or prediction processes and hold the potential to compromise the reliability and predictive accuracy of models. Consequently, we implement a sliding-window mechanism to mitigate coefficient jitter. Specifically, for the coefficients at time step $t$, we calculate new coefficients by applying a weighted average over a specified window, such as from $t-3$ to $t$. This method effectively reduces noise and temporal fluctuations in the time series.


\subsection{Results}
\label{Evaluations}
We conduct both quantitative and qualitative evaluations on the existing state-of-the-art methods, including SadTalker \cite{zhang2022sadtalker},  CodeTalker \cite{xing2023codetalker}, FaceFormer \cite{fan2022faceformer}, and MeshTalk \cite{richard2021meshtalk}. Not only comparing our method, VividTalker, with 3D facial animation methods, we also conduct comparison experiment on 2D facial animation method SadTalker \cite{zhang2022sadtalker} as it also disentangles facial animation into the head pose and mouth movement with two separate networks as the conflicted relationship of head pose and mouth movement. It's important to note that MeshTalk \cite{richard2021meshtalk}, Faceformer \cite{fan2022faceformer} and Codetalker \cite{xing2023codetalker} do not have head pose and detailed shape.


\begin{table*}[t]
\centering
\caption{User study using A/B testing and the percentage of preferences for A over B in the responses.}
\label{user_study}
\centering
\footnotesize
\begin{tabular}{llll}
\hline
    \textbf{Ours vs. Competitors}                          & \multicolumn{3}{c}{\textbf{Favorability}}        \\ \hline
 & \textbf{Naturalness} & \textbf{Synchronization} & \textbf{Average} \\ \hline
\textbf{Ours vs. MeshTalk \cite{richard2021meshtalk}}             & 95.3\%      & 93.2\%          & 94.3\% \\
\textbf{Ours vs. FaceFormer \cite{fan2022faceformer}}  & 86.1\%      & 87.5\%          & 86.5\%  \\
\textbf{Ours vs. CodeTalker \cite{xing2023codetalker}}  & 83.6\%      & 85.4\%          & 84.5\%  \\
\textbf{Ours vs. SadTalker \cite{zhang2022sadtalker}} & 84.7\%      & 82.5\%          &  83.6\%  \\
\textbf{Ours vs. RealSample} & 46.9\%      & 49.5\%          & 48.2\%  \\ \hline
\end{tabular}
\end{table*}

\begin{table*}[t]
\centering
\caption{Ablation study on Factor Disentanglement mechanism and Window Transformer. Dis. denotes the disentanglement mechanism. WinT denotes the Window Transformer.}
\footnotesize
    \begin{tabular}{lccccccc}
    \hline
                    & Pose Error$\downarrow$ & Mouth Error$\downarrow$ & Detail Error$\downarrow$ & FD$\downarrow$             & Diversity $\uparrow$      & LSE-C $\uparrow$         & LSE-D$\downarrow$           \\ \hline
Ours w/o Both  & 20.516                & 40.678        &  66.418                    &  0.103              &   0.695             & 0.502               &  14.244               \\
Ours w/o WinT. & 15.325                & 29.802        &  55.704                    & 0.061               &  1.004              &0.749              & 13.141               \\
Ours w/o Dis.  & 13.211          & 30.044               & 57.553               & 0.049          & 1.079         & 0.854          & 13.295          \\
Ours           & \textbf{8.852}  & \textbf{22.895}      & \textbf{46.984}      & \textbf{0.027} & \textbf{1.239} & \textbf{1.275} & \textbf{12.469} \\ \hline
    \end{tabular}    
\label{tab:disentangletab}
\end{table*}

\subsubsection{Quantitative Evaluation} 
\label{evaluation:Quantitative}
For quantitative evaluation, we demonstrate the superiority of our approach on multiple metrics that have been commonly used in previous studies: L2 error, Fréchet Distance (FD) \cite{frechet1957distance}, diversity \cite{zhang2022motiondiffuse}, LSE-D \cite{prajwal2020lip}, LSE-C \cite{prajwal2020lip}.

\textbf{Accuracy:} We employ L2 error to evaluate the accuracy between head pose, mouth movement, and detailed shape which are referenced as pose error, mouth error ,and detail error. The error is calculated by comparing the distance between the predictions and the real sample coefficients. We employ the Fréchet Distance (FD) \cite{frechet1957distance}, which is known as a measure of similarity between curves, to evaluate the distance between the predicted parameters and target parameters. 

\textbf{Diversity:} To evaluate the diversity of the generated head pose, the sequences are randomly divided into pairs, and the average collective distinctions are computed within each pair referring to the approach by MotionDiffuse \cite{zhang2022motiondiffuse}.

\textbf{Synchronization:} LSE-C and LSE-D are metrics proposed by wav2lip \cite{prajwal2020lip} to measure the lip-sync accuracy in videos. A higher mean confidence score, which is labeled as LSE-C, indicates a stronger alignment between audio and mouth movement. LSE-D involves calculating the average error by measuring the gap between the mouth and audio representations. A lower LSE-D indicates a stronger alignment between audio and mouth movement, meaning that speech and mouth movements are more synchronized.

As shown in Table~\ref{tab:comparison}, we can observe that our method, VividTalker, outperforms existing methods across a variety of evaluation metrics. This indicates that VividTalker achieves better performance in accuracy, head pose diversity and synchronization. Not only comparing to the previous 3D talking head animation methods, we also compare VividTalker with 2D method SadTalker \cite{zhang2022sadtalker}, a disentanglement-based method, in diversity, LSE-C, and LSE-D. We exclusively assess our performance against SadTalker using these criteria: diversity, LSE-C and LSE-D since the 3D facial model used in SadTalker \cite{zhang2022sadtalker} is different from ours.

Specifically, when we compared VividTalker with others in terms of L2 error, our method achieved up to a $59.8\%$ improvement in head pose, a $67.14\%$ improvement in mouth consistency, and a $43.5\%$ improvement in dynamic detail synchronization. On average, our method achieved a $54.9\%$ improvement across all L2 error metrics. Our method outperforms in Fréchet Distance referred as FD as well. The excellent results on L2 error and Fréchet Distance achieved by VividTalker suggest it can produce more accurate head pose, mouth movements, and detailed shape compared to the other methods. Besides, our method shows better head pose diversity as shown in Table \ref{tab:comparison}. We calculate the LSE-C and LSE-D over test videos and take the average for comparison. According to Table \ref{tab:comparison}, our method outperforms previous approaches by a large margin in LSE-C and LSE-D indicating that it generates more accurate mouth-synchronized movements. Even when compared to SadTalker which also disentangles facial animation into the head pose and mouth movement with two separate networks as the conflicted relationship of head pose and mouth movement, our method outperforms by $20.1\%$ on diversity and by $32.3\%$ on LSE-C.

\subsubsection{Qualitative Evaluation}

In this section, we present rendered results of state-of-the-art methods along with VividTalker on the word ``haven't'', aiming to perform a visual comparative analysis between them, as illustrated in Figure \ref{figcomparison}.  
The results reveal that the existing face animation techniques exhibit relatively diminished amplitudes of mouth movements and limited expressiveness. This deficiency can be attributed to the insufficient consideration of the coupling relationship between head and mouth movements. These methods tend to focus solely on one feature while disregarding the interactive effects between the two, consequently resulting in a restricted range of mouth motion variations. In contrast, our proposed approach successfully disentangles the distinct feature types, thereby enhancing the visual expressiveness in the rendered animations. Moreover, our method demonstrates superior fluency due to this disentanglement. By contrast, other methods occasionally encounter instances of static head movements or sudden changes in head pose, attributable to the model's inability to capture the intricate relationships between diverse feature types.

\subsubsection{User Study}
The degree of synchronization is readily discernible to the human visual and cognitive faculties, and the generation of stiffness is equally straightforward to capture. Hence, in addition to the quantitative and qualitative comparisons outlined above, we conducted a comprehensive user study to evaluate the performance of all the approaches as human evaluation offers a more reliable and direct assessment.
We extended invitations to 20 participants to engage in two distinct tasks: evaluating video clips generated by various methods and assessing them against the real sample. These tasks encompassed:
1) The assessment of video naturalness.
2) The comparison of mouth synchronization.
The collection comprised a total of $196$ video clips, and each row in Table \ref{user_study} details the evaluation of $196$ clip pairs, with each pair being spoken by a subject from the test set. Participants were tasked with selecting their preferred choice.
When pitted against baseline methods: MeshTalk \cite{richard2021meshtalk}, FaceFormer \cite{fan2022faceformer}, CodeTalker \cite{xing2023codetalker}, SadTalker \cite{zhang2022sadtalker}, our approach VividTalker, consistently outperformed in over $80\%$ of cases, excelling in both naturalness and mouth synchronization evaluations. 
The rendered results of SadTalkker \cite{zhang2022sadtalker} presents a more moderate and uniform head pose with little change. Besides, $83.6\%$ participants find that our method presents better naturalness and mouth synchronization than SadTalker \cite{zhang2022sadtalker}.
Impressively, favorability soared to $95.3\%$ when our method was juxtaposed with MeshTalk. Our approach stands as a formidable contender, despite only $46.9\%$ of the video clips surpassing the real sample.

\begin{figure*}[!t]
    \centering
    \includegraphics[width=0.9\textwidth]{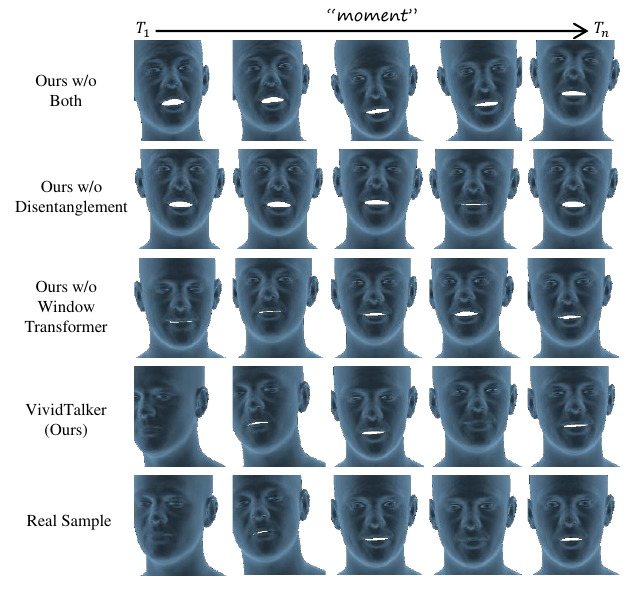}
    \caption{Qualitative results of ablation study on Factor Disentanglement and Window Transformer. Each row exhibits a head motion dynamic from $T_1$ to $T_n$. }
    \label{fig:ablation}
\end{figure*}

\begin{figure*}[t]
    \centering
    \includegraphics[width=0.9\textwidth]{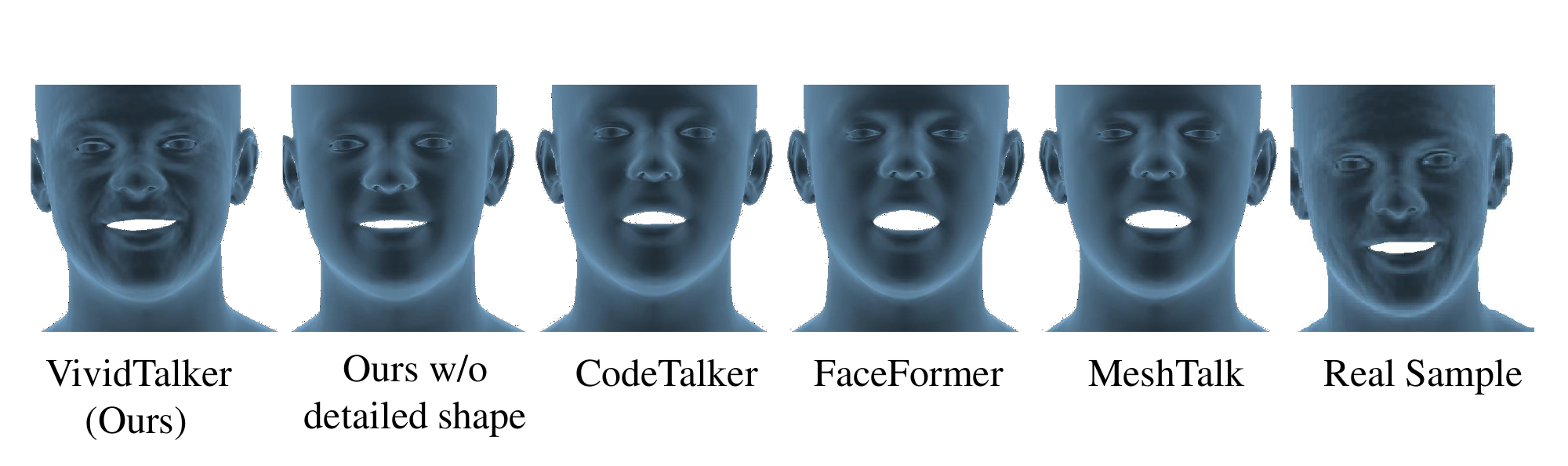}
    \caption{
   The rendering results with detailed shape become more vivid and better at conveying emotions.}
    \label{fig:details}
\end{figure*}

\begin{figure*}[t]
    \centering
    
    \includegraphics[width=0.9\textwidth]{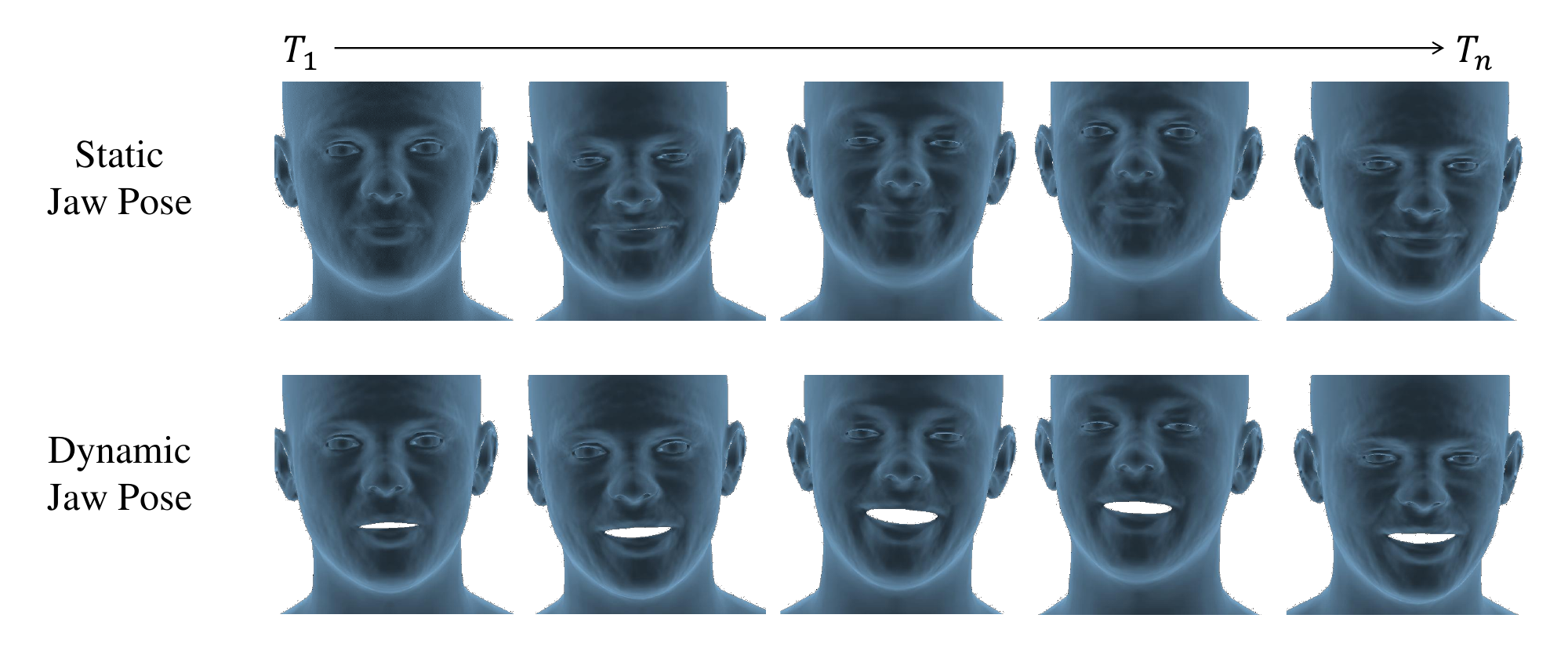}
    \caption{Visual comparison of rendered results with static jaw pose and dynamic jaw pose. Dynamic jaw pose is the Real Sample.}
    \label{fig:jaw}

\end{figure*}

\subsection{Ablation study}
We conduct an ablation study on Factor Disentanglement and Window Transformer mechanism from quantitative and qualitative evaluation. Similar to Section \ref{evaluation:Quantitative}, multiple metrics including L2 error, Fréchet distance \cite{frechet1957distance}, diversity \cite{zhang2022motiondiffuse}, LSE-D \cite{prajwal2020lip}, LSE-C \cite{prajwal2020lip} are employed to study the different components of our method. According to Table \ref{tab:disentangletab}, we observe that the disentanglement mechanism plays an important role in predicting accurate mouth movement. Figure \ref{fig:ablation} shows the visualization of the predicted sequence on the word ``moment''. When the disentanglement mechanism is removed, there is a significant difference between the mouth shape and the target mouth movement as shown in Figure \ref{fig:ablation}. This indicates that without factor disentanglement, the complex correlations between head movements and mouth movements can introduce confusion, impeding the model's ability to accurately learn distinct features. Moreover, as depicted in Figure \ref{fig:ablation}, the absence of disentanglement results in static rendered outputs and a tendency towards average mouth movements. 

Quantitatively (shown in Table \ref{tab:disentangletab}), the performance of various metrics has significantly declined following the elimination of the Window Transformer mechanism. Qualitatively, we find that the head pose and mouth movement start to deviate from the real sample. There is a plummet in both quantitative and qualitative results when eliminating disentanglement and the Window Transformer mechanism.

For the study on detailed shape, we exhibit the visualization with and without detailed shape as shown in Figure \ref{fig:details}. Texture details are an important attribute that defines the appearance of a face and can add realism to 3D models. In our experiments, we used a low-dimensional latent space of 128 dimensions to generate a detailed shape which is used to represent the wrinkles on the face. The expressiveness of the 3D facial model is significantly improved after adding details, allowing people to quickly and clearly perceive the emotional information conveyed by the 3D face as shown in Figure \ref{fig:details}. With the help of details, the vividness is been highly promoted.

\section{Analysis on Mouth Movements}
We split the pose coefficient $\theta$ into two parts as the first 3 dimensions are related to the head pose while the last 3 dimensions are related to mouth movement. The last 3 dimensions are designed to control the joint of the jaw. To clearly demonstrate the relationship between the last 3 dimensions and mouth movement, we conduct an experiment as shown in Figure \ref{fig:jaw}. The first row of  Figure \ref{fig:jaw} exhibits rendered results that we fix the last 3 dimensions. The second row of Figure \ref{fig:jaw} exhibits normal rendered results with dynamic changes in the last 3 dimensions. When the last 3 dimensions are fixed, the mouth movement stays in the same statement during the time stamp changes from $T_{1}$ to $T_{n}$ as shown in Figure \ref{fig:jaw}. When the last 3 dimensions change normally, the mouth movement of the rendered results changes naturally. The rendered results demonstrate that it is reasonable to use the first 3 dimensions for head pose control and the last 3 dimensions for mouth movement control.

\section{Conclusions}

In conclusion, we introduce VividTalker, an innovative framework for speech-driven 3D head animation that overcomes the challenges of factor disentanglement and detail enrichment without relying on real sample scanned 3D scanned mesh. By utilizing separate encodings for head pose and mouth movement using VQ-VAE models, our framework effectively resolves feature learning conflicts, resulting in more accurate and precise 3D head animation. Additionally, leveraging a pre-trained DECA model enables us to enrich the animation with dynamic detailed shapes, enhancing the visual fidelity. Furthermore, we provide a comprehensive 3D dataset constructed with a pre-trained 3D reconstruction model, serving as a valuable resource for 3D head animation training and evaluation.

\textbf{Data Availability} The data used in the experiments are publicly available online via \href{https://weizhaomolecules.github.io/VividTalker/}{https://weizhaomolecules.github.io/VividTalker/}. 

\textbf{Code Availability} Codes for VividTalker and comparison videos  are available on \href{https://weizhaomolecules.github.io/VividTalker/}{https://weizhaomolecules.github.io/VividTalker/}.






\backmatter






\bibliographystyle{sn-mathphys} 
\bibliography{ref}

\end{document}